\documentclass[letterpaper, 10 pt, conference]{ieeeconf}  

\IEEEoverridecommandlockouts                              

\usepackage[dvipdfmx]{graphicx}
\usepackage{epsfig} 
\usepackage{mathptmx} 
\usepackage{times} 
\usepackage{amsmath} 
\usepackage{amssymb}  

\usepackage{bm}
\usepackage{subcaption} 
\usepackage{caption}
\usepackage{multirow} 
\usepackage{subcaption} 
\usepackage{comment}

\usepackage{algorithm}
\usepackage{algorithmicx}
\usepackage[noend]{algpseudocode}
\usepackage{color}
\usepackage{xcolor}
\usepackage{url}
\usepackage{caption}

\DeclareMathAlphabet{\mathcal}{OMS}{cmsy}{m}{n}

\title{\LARGE \bf
Visual-Based Forklift Learning System Enabling Zero-Shot Sim2Real Without Real-World Data
}

\author{Koshi Oishi$^{*,1}$, Teruki Kato$^{1}$, Hiroya Makino$^{1}$, and Seigo Ito$^{1}$ 
\thanks{$^{1}$ The authors are with Toyota Central R\&D Labs., Inc., 41-1 Yokomichi, Nagakute, Aichi, Japan.}
        \thanks{$^{*}$ Corresponding author. {\tt\small e1616@mosk.tytlabs.co.jp}}%
        }

\begin{document}

\maketitle
\thispagestyle{empty}
\pagestyle{empty}

\begingroup
  \renewcommand\thefootnote{}            
  \footnotetext{\footnotesize
  © 2025 IEEE. Personal use of this material is permitted. 
  Permission from IEEE must be obtained for all other uses, in any current or future media, 
  including reprinting/republishing this material for advertising or promotional purposes, 
  creating new collective works, for resale or redistribution to servers or lists, 
  or reuse of any copyrighted component of this work in other works.}
  \addtocounter{footnote}{-1}            
\endgroup

\begin{abstract}
Forklifts are used extensively in various industrial settings and are in high demand for automation. 
In particular, counterbalance forklifts are highly versatile and are employed in diverse scenarios. 
However, efforts to automate these processes are lacking, primarily owing to the absence of a safe and performance-verifiable development environment. 
This study proposes a learning system that combines a photorealistic digital learning environment with a 1/14-scale robotic forklift environment to address this challenge.
Inspired by the training-based learning approach adopted by forklift operators, we employ an end-to-end vision-based deep reinforcement learning approach.
The learning is conducted in a digitalized environment created from CAD data, making it safe and eliminating the need for real-world data. 
In addition, we safely validate the method in a physical setting using a 1/14-scale robotic forklift with a configuration similar to that of a real forklift. 
We achieved a 60\% success rate in pallet loading tasks in real experiments using a robotic forklift.
Our approach demonstrates zero-shot sim2real with a simple method that does not require heuristic additions. 
This learning-based approach is considered a first step towards the automation of counterbalance forklifts.

\end{abstract}


\section{Introduction} \label{sec:intro}
Forklifts are essential in various industries, including factories, logistics centers, ports, and construction sites. 
In particular, counterbalance forklifts are central to the industry owing to their robustness and power, generating significant demand for automation.
Recent advancements have led to the automation of reach-type forklifts, considering their low power requirements and high maneuverability \cite{LindeAutomation2024}.
In contrast, automating counterbalance forklift operations requires advanced controllers that can leverage the versatility and overcome the limited maneuverability of forklifts.
As humans learn these tasks through training, applying deep reinforcement learning (DRL) to automation is a natural progression \cite{end2end_human, levineEnd, kaufmann2023champion}. 
However, research on its application to forklifts remains limited,
likely owing to the risks of conducting experiments in real environments and the lack of training datasets.
Therefore, a learning system that can overcome these challenges is required.

DRL has issues such as the dangers of real-world training and low sample efficiency.
Sim2real, in which a policy is trained in a digital environment and transferred to the real world, is a common strategy for these issues \cite{RN33}.
However, learning in a digital environment creates a new challenge known as the domain gap.
Various methods have been proposed to bridge this gap. 
However, as collecting failure data poses significant challenges, methods that do not rely on real-world data are preferable for forklifts.
Notable studies without real-world data include those focusing on soccer and mountain climbing with bipedal and quadrupedal robots, respectively \cite{RN31, RN22}.  
These studies used domain randomization, which randomizes the environment during training \cite{RN46, Abeel_DR}.
However, the use of visual information in these studies was limited.
Moreover, as opposed to these small-scale robots, the first validation of a sim2real method using large machines such as forklifts carries risks, even though safety is verified in the digital environment.

\begin{figure}[t]
	\centering
		\includegraphics[keepaspectratio, width=8.5cm]{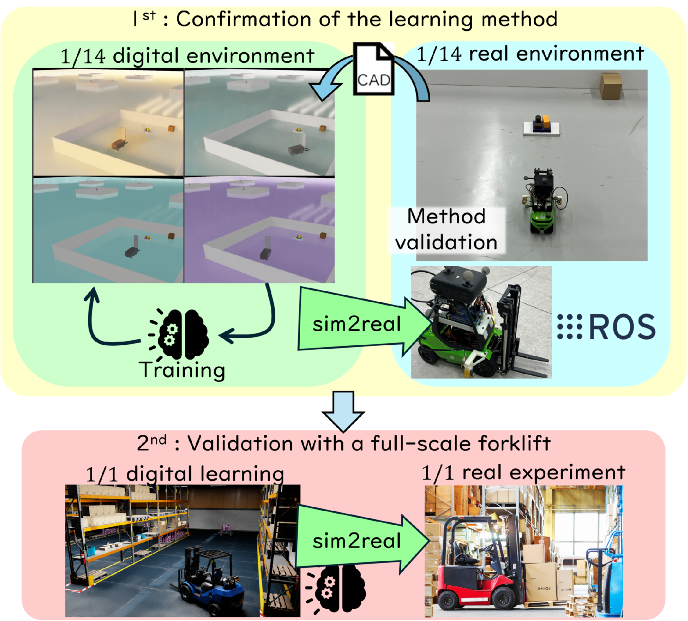}
	  \caption{Proposed concept. To implement DRL-based control on a full-scale forklift, we first test sim2real in a smaller environment. Sim2real is applied to the full-scale forklift after confirming the safety of the method.}\label{fig:zu1}
\end{figure}

We propose a practical learning system for automating counterbalance forklifts in pallet loading tasks involving the pallet approach and loading decision.
In the pallet approach, a policy is optimized using end-to-end DRL to control the forklift based on the same visual and velocity information as that of humans operators.
A loading decision policy to perform a loading task is acquired through supervised learning.
Our concept involves training and validation in 1/14-scale digital and real environments and subsequent extension of the learning method to full scale after confirming the safety in these environments, as shown in Fig.~\ref{fig:zu1}.
Furthermore, to address the domain gap in which real-world data is lacking, we use domain randomization with NVIDIA Omniverse Isaac Sim (Isaac Sim), which is a highly advanced photorealistic digital environment \cite{RN49}.
Constructing this digital environment requires only CAD data, thereby eliminating the need for real-world data in training. 
Moreover, the 1/14-scale forklift used in the real environment features front-wheel drive and rear-wheel steering, similar to the full-scale version, and its lift function operates hydraulically. 
We developed the 1/14-scale forklift using a robot operating system (ROS) \cite{RN64}. 
This study represents the first step towards the learning-based control of real counterbalance forklifts.
The contributions of our research are as follows:

\begin{itemize}
    \item We propose a forklift learning system using a photorealistic digital environment that allows for safe research and validation.
    \item We introduce an end-to-end DRL method for vision-based forklift control that focuses on the pallet approach, which does not require real-world data.
    \item We develop a dataset construction method for the pallet loading decision policy using a photorealistic digital environment.
    \item We demonstrate zero-shot sim2real using a 1/14-scale forklift.
\end{itemize}

The remainder of this paper is organized as follows. 
Related works are introduced in Section \ref{sec:work}.
Section \ref{sec:RL} presents the target tasks and learning setup for the proposed system.
Our proposed learning system that uses a photorealistic simulator is described in Section \ref{sec:sysfork}.
Section \ref{sec:demo} outlines the demonstration in the real environment.
Finally, concluding remarks and future directions are presented in Section \ref{sec:demo}.

\section{Related work} \label{sec:work}
Research on the automation of counterbalance forklifts has been limited to date. 
Walter et al. \cite{RN65} and Teller et al. \cite{RN34} focused on non-learning-based methods that use multiple LiDAR sensors for environmental perception. 
However, to the best of our knowledge, few studies have explored real-world implementations of the automation of counterbalance forklifts.
Hadwiger et al. \cite{RN32} conducted DRL using visual inputs on a counterbalance forklift in a digital environment. 
However, their method did not address the domain gap issue sufficiently and was only evaluated digitally.

End-to-end DRL, which acquires policies through training in a manner similar to human learning, can achieve human-like performance \cite{end2end_human,levineEnd}.
Trained human operators play a central role in operating counterbalance forklifts, contributing to the adoption of the end-to-end DRL method. 
Common tasks in DRL applications include navigation and dexterous manipulation \cite{RN33, D4RL}.
Forklifts require both navigation and dexterous operation because the forks are directly attached to the wheels. 
In previous research, such tasks were addressed by using a robotic car for imprecise navigation and a robotic arm for dexterous manipulation, whereas our task requires achieving dexterous navigation solely through the wheels \cite{RN35}.

DRL has been studied extensively in sim2real scenarios because of the dangers associated with real-world experimentation and the low sample efficiency of such experiments \cite{RN33, RN46, RN28, RN58}.
Domain randomization is an effective method for addressing the domain gap in sim2real without relying on real-world data, and it has been widely used in many studies \cite{RN46,RN43,RN45, james2019sim}.
Among these, the research closest to ours is the study on soccer, in which both navigation and dexterous manipulation were required \cite{RN31}. 
Similar to our method, this research progressed to the stage of performing soccer tasks using low-cost onboard cameras \cite{soccer_new}.
However, NeRF was employed to achieve vision-based tasks, thereby negating the advantage of not using real-world data, which was a key strength in previous studies \cite{Nerf}.
Furthermore, although NeRF is effective for static visuals, it is not well suited to moving objects such as pallets or loads \cite{Byravan}.

Recent advancements have made it possible to create photorealistic digital environments in which human training can be performed using VR headsets \cite{isaacSim}. 
Mittal et al. \cite{RN49} and Yu et al. \cite{RN60} proposed photorealistic learning systems using Isaac Sim. 
Although these studies used photorealistic digital environments to generate datasets for supervised learning, they did not report on vision-based DRL. 
Our study leverages the photorealism of Isaac Sim and domain randomization to address the zero-shot sim2real of vision-based DRL, which does not require real-world data and can adapt to dynamic environments.

\section{Vision-based forklift control via DRL} \label{sec:RL}
This section first describes the targeted tasks, followed by the learning setup employed by our learning system.
The goal is to develop a forklift controller that can perform forklift tasks. 
This study emphasizes the overall learning framework, including the practical implementation; therefore, we use simple methods for training.

\subsection{Target task} \label{sec:task}
The target task is the pallet loading operation by a forklift. 
It involves randomly positioning the forklift within the visible range of the pallet and then executing the loading process. 
We divide this task into two phases: the approach to the pallet and the lifting operation.
Different policies are applied to each phase.
We use a simple decision policy to determine whether to lift the pallet. 
Therefore, the focus is on the learning method related to the forklift-specific approach to the pallet.
The observation data used for this task include visual and velocity information, similar to that used by a human operator. 
In addition, forklift operators lean to the sides to check the alignment between the forks and pallets. 
Therefore, we installed two cameras on the left and right sides of the forklift to capture images of the forks and pallets. 

\subsection{Policy design for pallet approach} \label{sec:poli}
The approach policy is designed to output the throttle and steering of the forklift based on the visual and velocity inputs obtained from the camera images and velocity data, respectively.
The camera images, which are 352 $\times$ 288 RGB images, are first resized to 224 $\times$ 224.
These resized images are then converted into feature vectors, $L_\mathrm{l} \in \mathbb{R}^{512}$ and $L_\mathrm{r} \in \mathbb{R}^{512}$, using a ResNet pretrained on ImageNet \cite{Resnet, imagenet}.
Subsequently, these feature vectors are used as inputs for the policy.
This method was proposed as Resnet as representation for reinforcement learning \cite{RRL}. 
No special processing is required because we use a standard pretrained ResNet without domain adaptation. 
In addition, assuming that speed sensors are installed, we utilize the two-dimensional speed information $\bm{v} := [v_{\mathrm x} , v_\mathrm{y}]^{\top}$ and yaw rate $\omega$.
Finally, we include the past actions $a_{\mathrm{old1}}$ and $a_{\mathrm{old2}}$. 
Thus, the observation space $\mathcal{O}$ is a 1029-dimensional vector space.

\begin{figure}[t]
\vspace*{1.4mm}
	\centering
		\includegraphics[keepaspectratio, width=8.5cm]{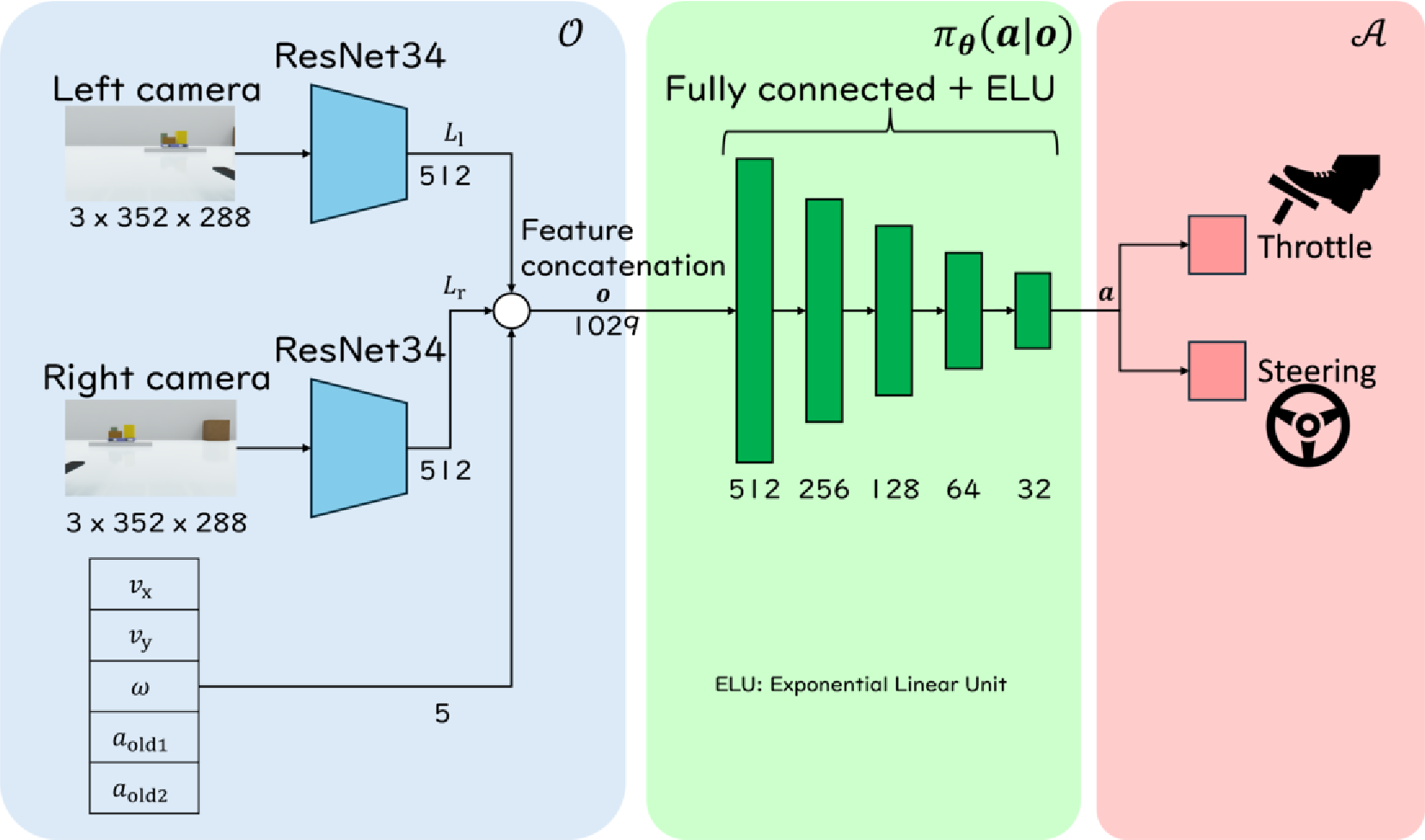}
	  \caption{Approach policy design.}\label{fig:poli}
\end{figure}

The policy network $\pi_{\bm{\theta}} (\bm{a} | \bm{o})$ outputs the normalized throttle and steering commands for the forklift. 
Here, $\bm{a} \in \mathcal{A} = [-1 ,1]^2$ represents the actions and $\bm{\theta}$ denotes the parameters of the policy network. 
The network $\pi_{\bm{\theta}} (\bm{a} | \bm{o})$ consists of five fully connected layers with exponential linear units as activation functions. 
The overall structure is shown in Fig.~\ref{fig:poli}.

\subsection{Training method} \label{sec:train}
We use the on-policy reinforcement learning algorithm proximal policy optimization (PPO) to train the approach policy $\pi_{\bm{\theta}} (\bm{a} | \bm{o})$ \cite{RN50}. 
The following subsections describe the loss and reward functions employed in our method.

\subsubsection{Loss function}
The actor and critic networks share a common network, which is optimized using the following single loss function:
\begin{align}
    \begin{split}
        L_t(\bm{\theta}) &= \mathbb{E}_t \left[L_t^{\mathrm{PPO}} (\bm{\theta}) + c_1 L_t^{\mathrm{value}}(\bm{\theta}) - c_2 H \left(\pi_{\bm{\theta}}(\cdot | \bm{o}_t) \right) \right. \\
        & \left. \hspace{10mm} + c_3 L_t^{\mathrm{bound}}(\bm{\theta}) \right] ,
    \end{split}
\end{align}
where $L_t^{\mathrm{PPO}} (\bm{\theta})$ denotes the loss of the PPO policy, $L_t^{\mathrm{value}}(\bm{\theta}) $ denotes the state-value loss, $H_t \left(\pi_{\bm{\theta}}(\cdot | \bm{o}_t) \right)$ denotes the entropy, $ L_t^{\mathrm{bound}}(\bm{\theta})$ denotes the boundary loss, and $c$ denotes the weight.
Moreover, $L_t^{\mathrm{PPO}}(\bm{\theta})$ is determined by
\begin{align}
    \begin{split}
L_t^{\mathrm{PPO}}(\bm{\theta}) &= -\mathbb{E}_t \left[ \min \left(\rho_{t}(\bm{\theta}) \hat{A}_t, \text{clip}(\rho_{t}(\bm{\theta}), 1-\epsilon, 1+\epsilon)\hat{A}_t \right) \right], \\
\rho_{t}(\bm{\theta}) &= \frac{\pi_{\bm{\theta}}(\bm{a}_t | \bm{o}_t)}{\pi_{\bm{\theta}_{\text{old}}}(\bm{a}_t | \bm{o}_t)},
    \end{split}
\end{align}
where the clip function smooths the gradients by limiting the loss changes and $\epsilon$ denotes the clipping parameter.
In addition, $\hat{A}_t$ is computed using generalized advantage estimation \cite{RN50}.
Furthermore, $L_t^{\mathrm{value}}(\bm{\theta})$ is obtained by
\begin{align}
L_t^{\mathrm{value}}(\bm{\theta}) &= \mathbb{E}_t \left[ \left(V_{\bm{\theta}}(\bm{s}_t) - V^{\text{target}}_t \right)^2 \right],
\end{align}
where $V_{\bm{\theta}}(\bm{s}_t)$ denotes the state-value function, $V^{\text{target}}_t$ denotes the target state-value function, and $\bm{s}_t$ denotes the state of the environment, including privileged information.
In addition, entropy is introduced to encourage exploration, as follows:
\begin{align}
H(\pi_{\bm{\theta}}(\cdot | \bm{o}_t)) &= -\sum_{\bm{a}} \pi_{\bm{\theta}}(\bm{a} | \bm{o}_t) \log \pi_{\bm{\theta}}(\bm{a} | \bm{o}_t).
\end{align}
Finally, the boundary loss $L_{\text{bound}}$ is introduced to prevent the actions from taking excessively large values: 
\begin{align}
     L_t^{\mathrm{bound}}(\bm{\theta}) = \|\bm{\mu}(\bm{o}_t) \|,
\end{align}
where $\bm{\mu}(\bm{o}_t)$ denotes the mean vector of the actions output by the policy $\pi_{\bm{\theta}}( \cdot | \bm{o}_t)$.

\subsubsection{Reward function} \label{sec:rw}
\begin{figure}[t]
\vspace*{1.4mm}
	\centering
		\includegraphics[keepaspectratio, width=7cm]{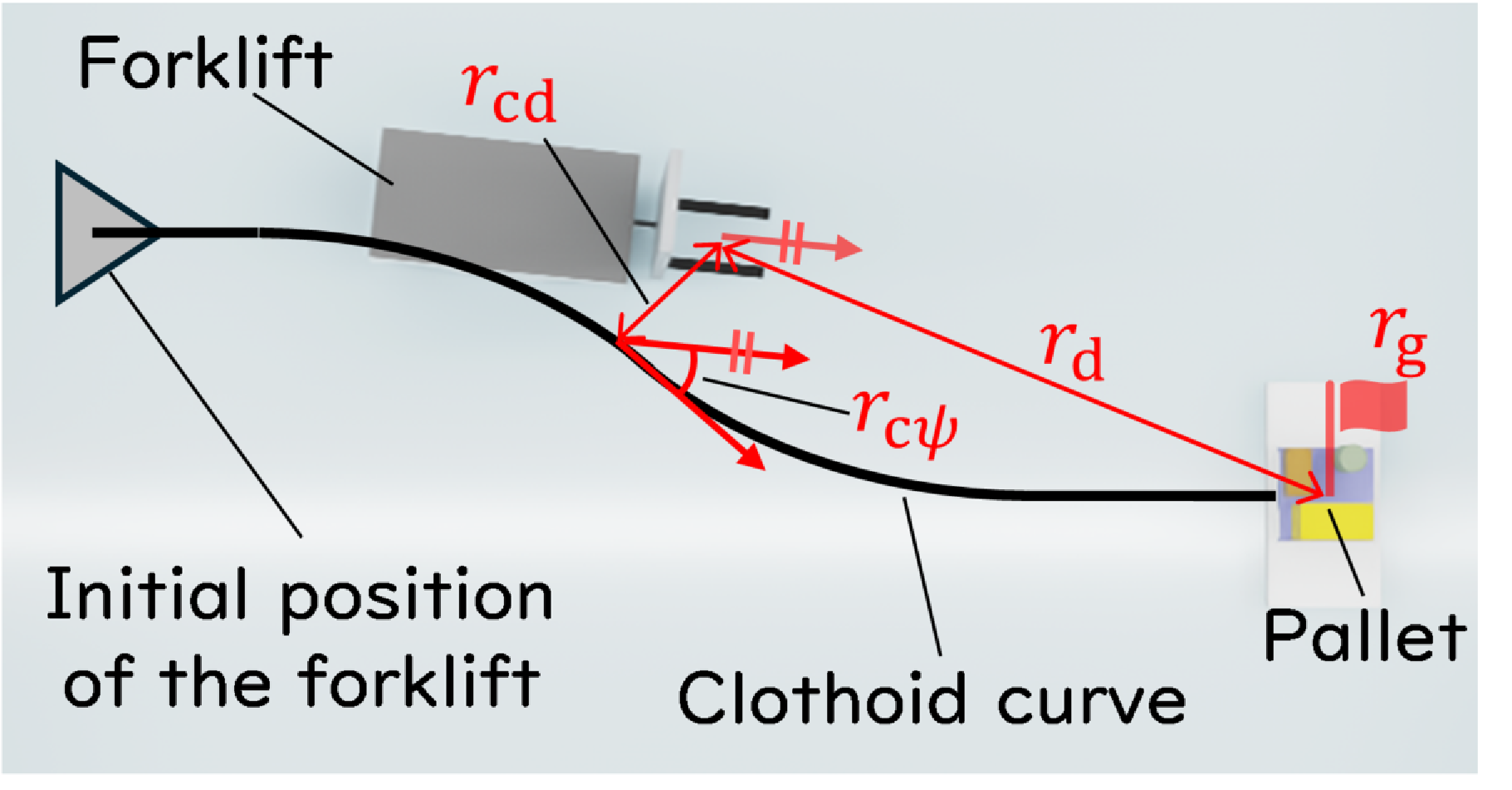}
	  \caption{State used for calculating $R_{+}$.}\label{fig:rw}
\end{figure}

Our reward function consists of positive and penalty rewards for desirable and undesirable states, respectively.
In addition, calculating these rewards uses states of privileged information that are available in the digital environment.
The deviation from a reference trajectory is used for the positive rewards. 
The initial position is the location of the forklift at the start of the task, and the terminal position is the pallet.
The reference trajectory is based on the approximation of a clothoid curve, which remains fixed throughout the task. 
The positive reward $R_{+}$ is defined as follows:
\begin{align}
     R_{+} = \alpha_1 \frac{1}{r_{\mathrm d}} + \alpha_2 \frac{1}{r_{\mathrm{cd}}} + \alpha_3 \frac{1}{r_{\mathrm{c}\psi}} +\alpha_4 r_{\mathrm g},
\end{align}
where $r_{\mathrm d}$ and $r_{\mathrm{cd}}$ are the distances from the center of the forks to the pallet and clothoid curve, respectively, $r_{\mathrm c \psi}$ is the difference between the orientation of the forks and the tangent to the clothoid curve, $r_{\mathrm g} \in \{0,1\}$ is a special reward that is assigned when the forks reach the pallet position, and $\alpha$ represents the weight.
The components of $R_{+}$ are illustrated in Fig.~\ref{fig:rw}.

As opposed to general navigation tasks, our task requires inserting the forks without moving the pallet.
Therefore, the penalty rewards are defined as follows:
\begin{align}
     R_{-} = \alpha_5 r_{\mathrm p} + \alpha_6 r_{\mathrm v} + \alpha_7 r_{\mathrm a}  + \alpha_8 r_{\mathrm{ini}} ,
\end{align}
where
\begin{align*}
     r_{\mathrm p} &= \left\{
          \begin{array}{ll}
               -1 & \textbf{if } \|\bm{v}_{\mathrm p} \| > 0.01 ,\\
               0 & \textbf{otherwise} ,
          \end{array} \right. \\
     r_{\mathrm v} &= \left\{
          \begin{array}{ll}
                -(\|\bm{v}\| -0.07)^2 & \textbf{ if } \|\bm{v}\| > 0.07 ,\\
               0 & \textbf{otherwise} ,
          \end{array} \right. \\
     r_{\mathrm a} &= - \|\bm{a} - \bm{a}_{\mathrm{old}} \|^2 ,\\
     r_{\mathrm{ini}} &= \left\{
          \begin{array}{ll}
               -1 & \textbf{if } \|\bm{v}\| < 0.05 \textbf{ and } r_{\mathrm d} > 0.3 ,\\
               0 & \textbf{otherwise} ,
          \end{array} \right. 
\end{align*}
where $\bm{v}_{\mathrm p}$ is the velocity of the pallet and $\bm{a}_{\mathrm{old}}$ is the previous action. 
$r_{\mathrm p}$ is the penalty for contacting the pallet; a negative reward is assigned if the pallet starts to move. 
$r_{\mathrm v}$ and $r_{\mathrm a}$ are penalties that are designed to bridge the gap between the digital and real environments. 
Specifically, $r_{\mathrm v}$ prevents excessive speed, whereas $r_{\mathrm a}$ prevents the actions from becoming erratic. 
$r_{\mathrm{ini}}$ is a penalty to prevent the forklift from freezing at the initial position.
Therefore, the reward function can be summarized as follows:
\begin{align}
     R = R_{+} + R_{-}.
\end{align}

\section{Forklift learning system} \label{sec:sysfork}
Learning-based controls often have black-box characteristics, which makes it challenging to validate them in high-powered, contact-heavy industrial machinery. 
Therefore, we constructed digital and real environments on a 1/14 scale. 
The digital environment was created using Isaac Sim, which supports the development of photorealistic environments \cite{RN49}. 
We leveraged the extensive randomization capabilities and parallel learning features of the tool to address the domain gap challenges of reinforcement learning. 
The real environment was constructed using a ROS \cite{RN64}.

\begin{figure*}[ht]
\vspace*{1.4mm}
    \centering
        \begin{minipage}[b]{0.35\textwidth}
            \centering
            \includegraphics[width=6cm]{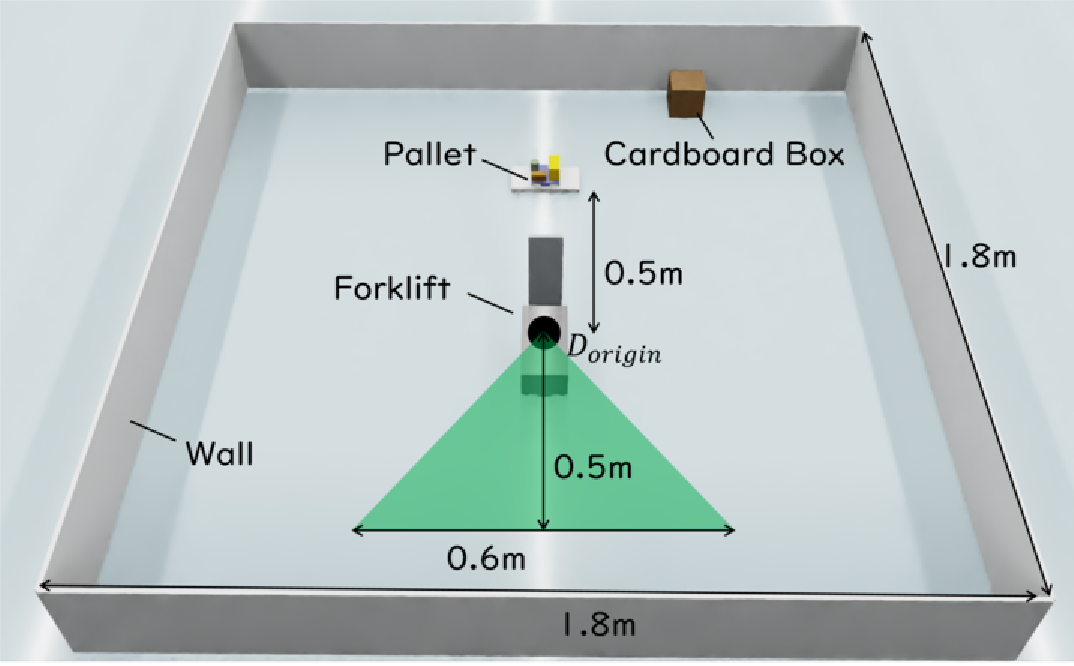}
            \subcaption{Digital environment}\label{fig:digi_env}
        \end{minipage} 
        \begin{minipage}[b]{0.3\textwidth}
            \centering
            \includegraphics[width=5.1cm]{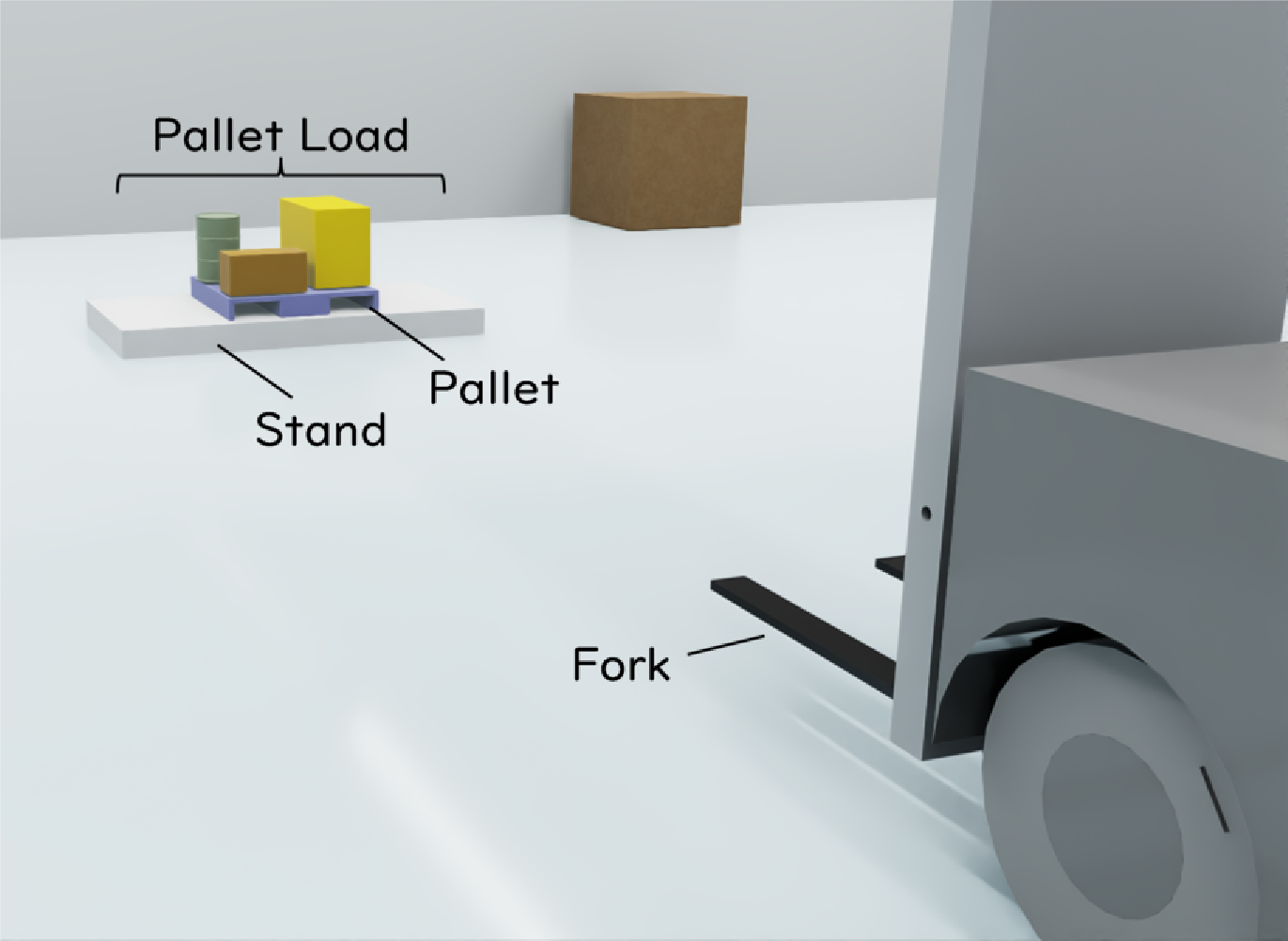}
            \subcaption{Side view (digital)}
        \end{minipage} 
        \begin{minipage}[b]{0.28\textwidth}
            \centering
            \includegraphics[width=5.2cm, trim= 0cm 0.0cm 0cm 1.3cm,clip]{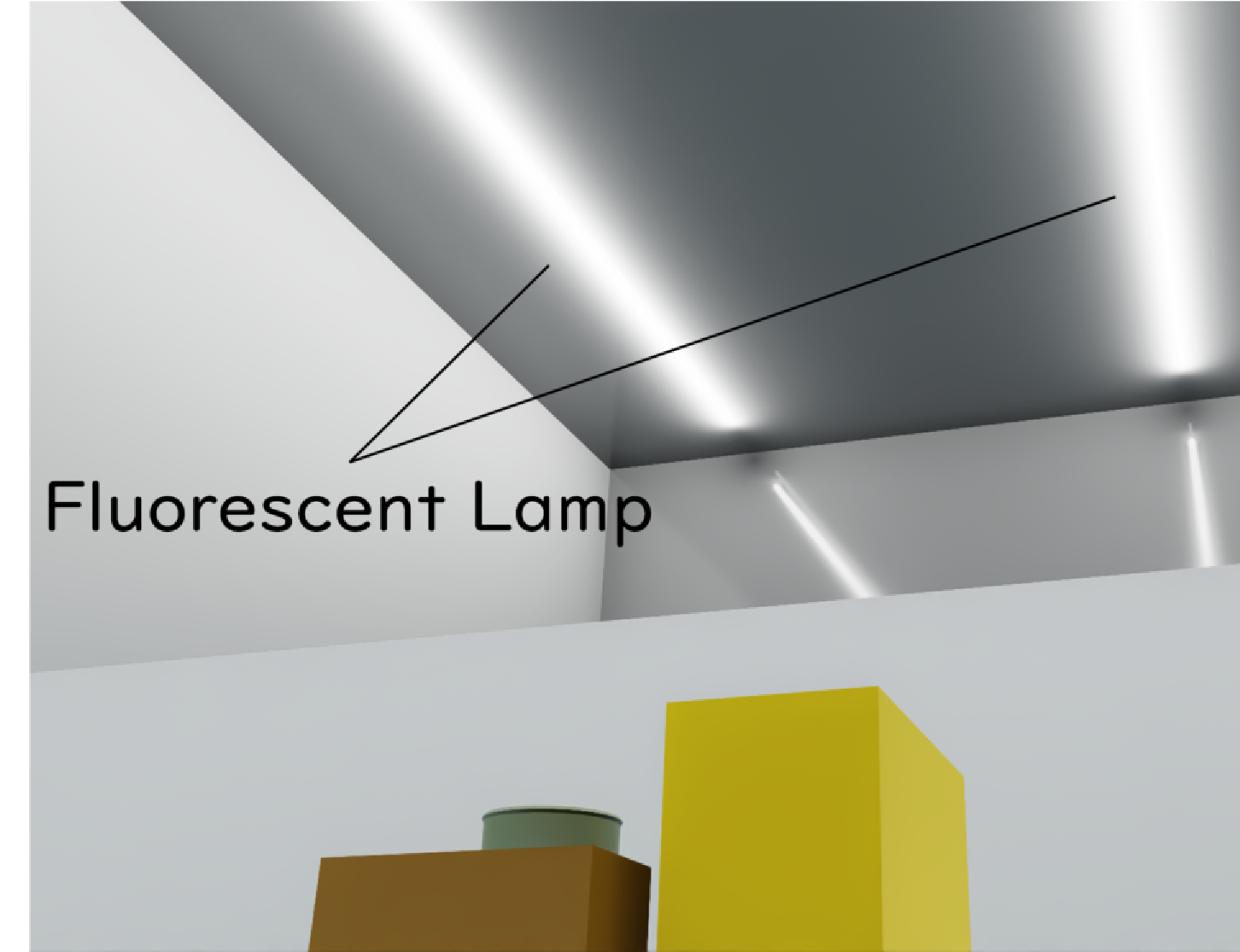}
            \subcaption{Ceiling (digital)}\label{fig:ceiling}
        \end{minipage} \\
        \begin{minipage}[b]{0.35\textwidth}
            \centering
            \includegraphics[width=6cm]{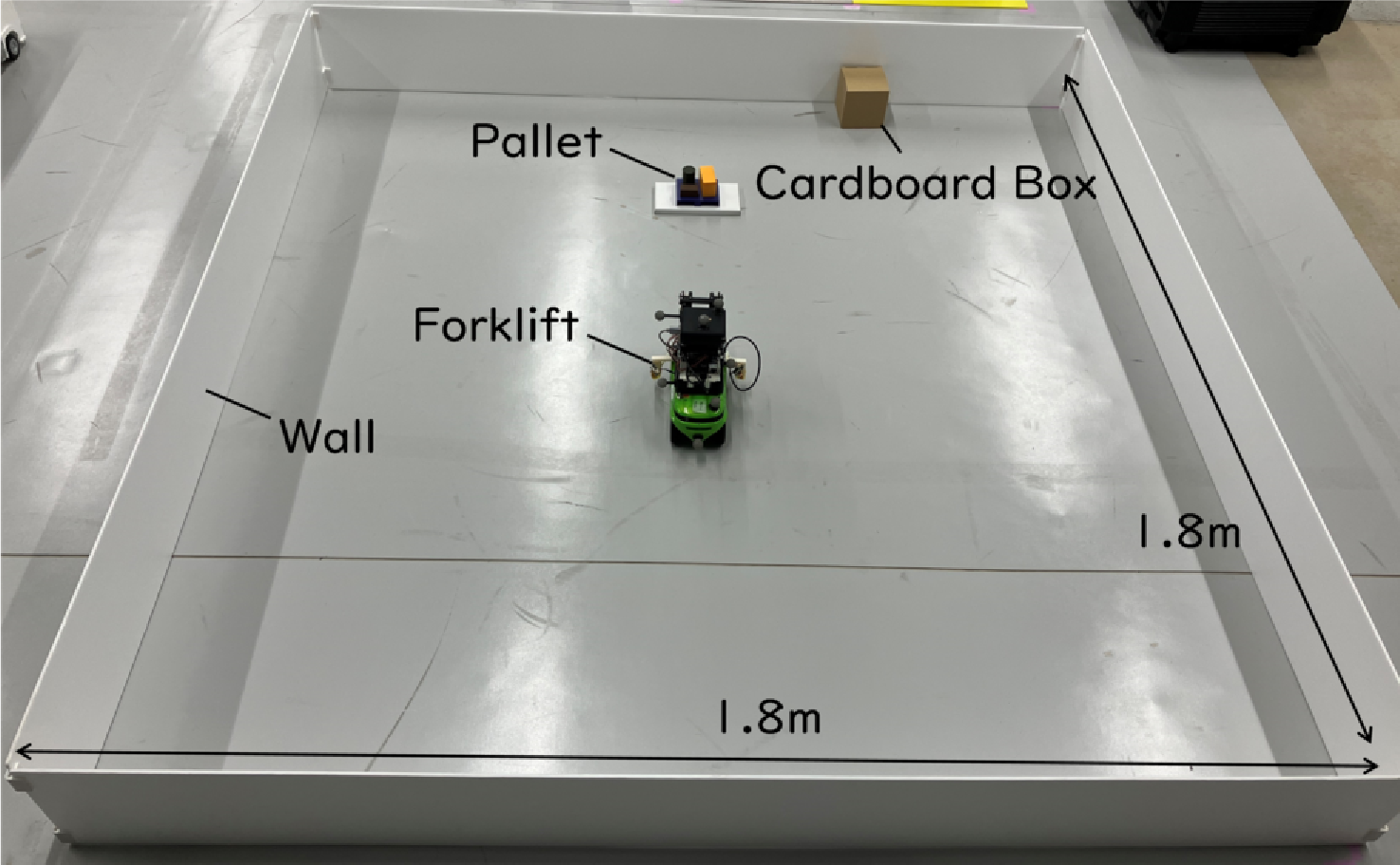}
            \subcaption{Real environment}
        \end{minipage} 
        \begin{minipage}[b]{0.3\textwidth}
            \centering
            \includegraphics[width=5.1cm, trim= 0cm 0.7cm 0cm 0.1cm,clip]{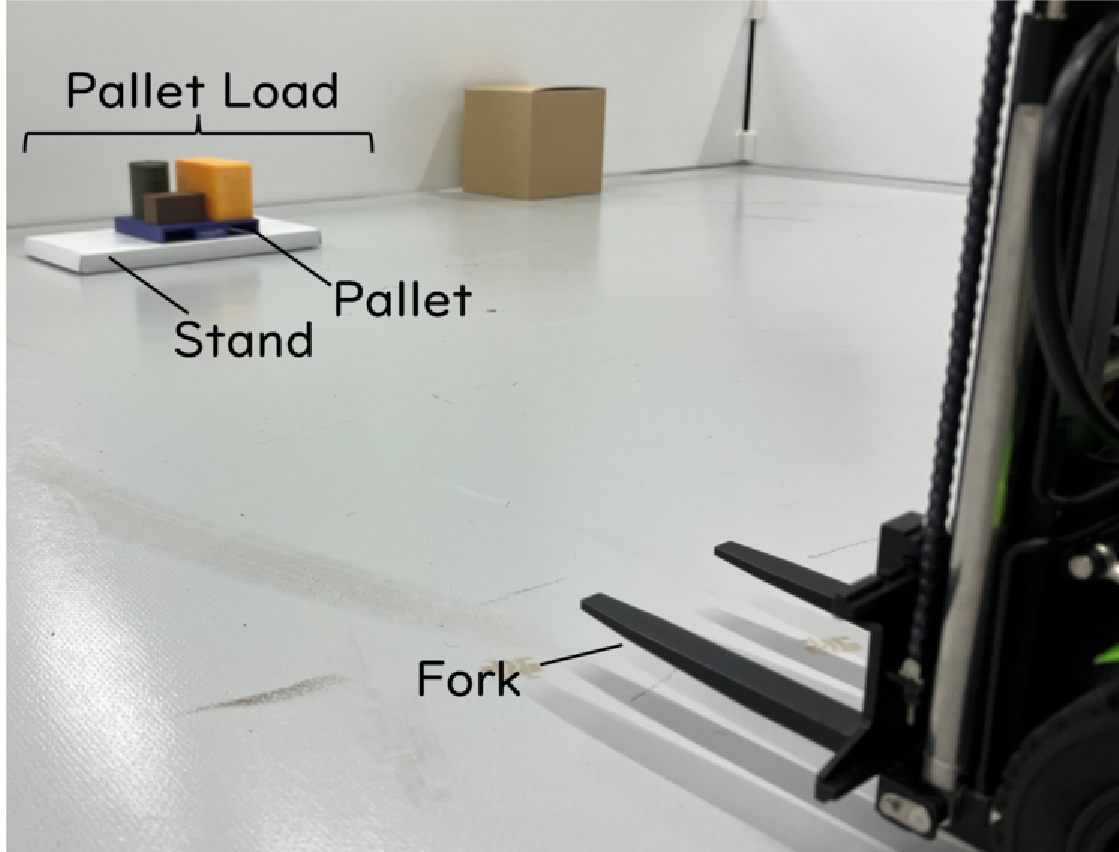}
            \subcaption{Side view (real)}
        \end{minipage} 
        \begin{minipage}[b]{0.28\textwidth}
            \centering
            \includegraphics[width=5.2cm]{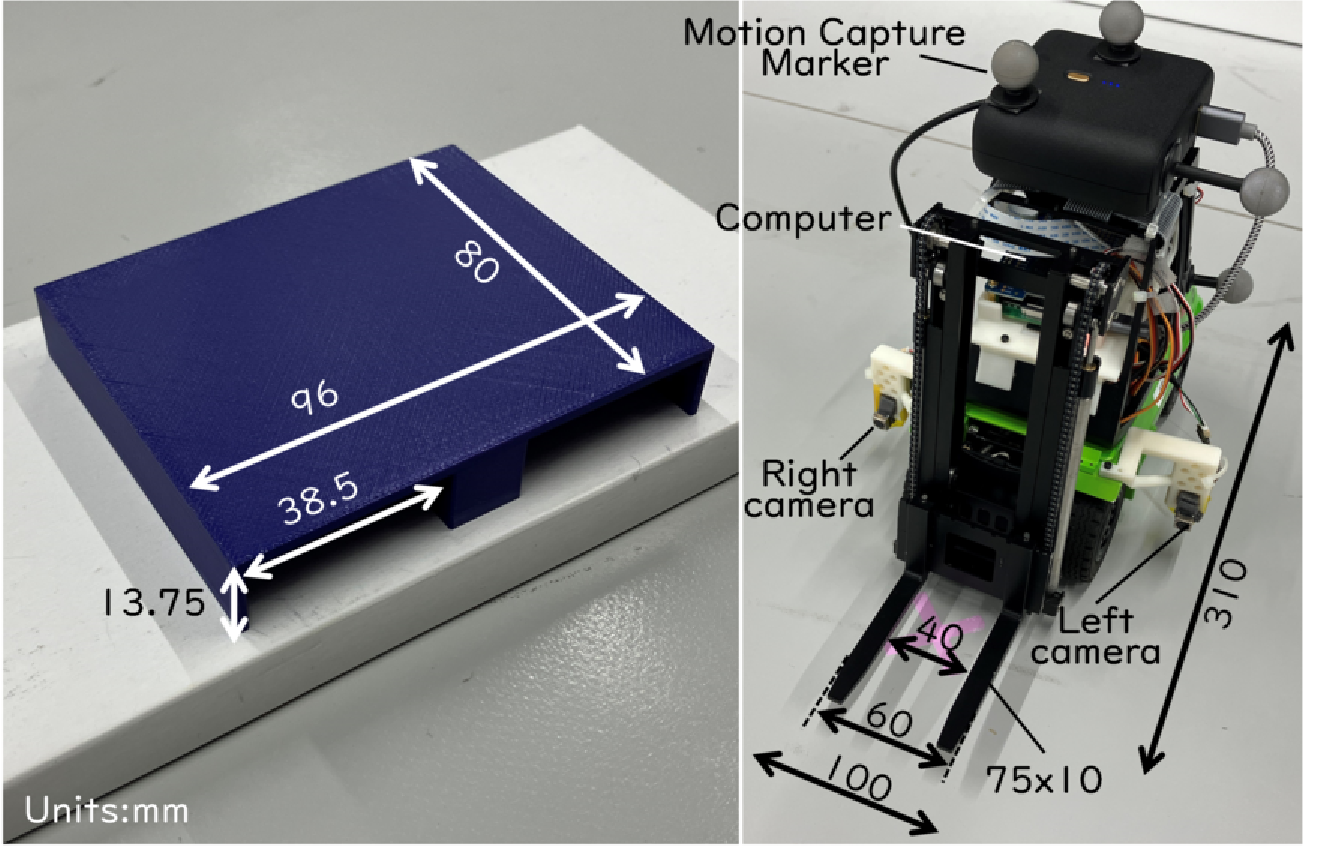}
            \subcaption{Sizes of forklift and pallet}\label{fig:fork_pallet}
        \end{minipage} \\
    \caption{Digital and real environments. (a), (b), and (c) represent the digital environment. The green triangle in (a) indicates the initial position of the forklift, which is randomly determined at the start of the task. $D_{origin}$ denotes the origin. (d) and (e) represent the real environment. (f) represents the size of the real forklift and pallet. Units are in mm.}\label{fig:twin}
\end{figure*}

\begin{table}[t]
     \centering
     \caption{Domain randomization targets}
     \label{tb:dr}
     \begin{tabular}{|c|c|c|} \hline
          Item & Range  \\ \hline
          Observed speed & $\pm 10\%$ of value \\ \hline
          Action & $\pm 10\%$ of command \\ \hline
          Floor color & $\pm 20\%$ RGB \\ \hline
          Pallet stand color & $\pm 20\%$ RGB \\ \hline
          Pallet and load color & $\pm 20\%$ RGB \\ \hline
          Light intensity & $100$--$100000$ lm \\ \hline
          Light temp. & $2000$--$7500$ K \\ \hline
     \end{tabular}
\end{table}

\begin{figure}[t]
     \begin{minipage}[b]{0.45\linewidth}
         \centering
         \includegraphics[keepaspectratio, scale=0.5, trim= 0cm 0.2cm 0cm 0.6cm,clip]{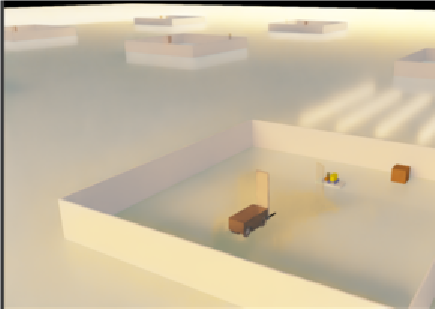}
         \subcaption{50564 lm, 5407 K, floor RGB (0.48, 0.59, 0.56)}\label{fig:dr_1}
     \end{minipage} 
     \begin{minipage}[b]{0.45\linewidth}
         \centering
         \includegraphics[keepaspectratio, scale=0.5, trim= 0cm 0.2cm 0cm 0.6cm,clip]{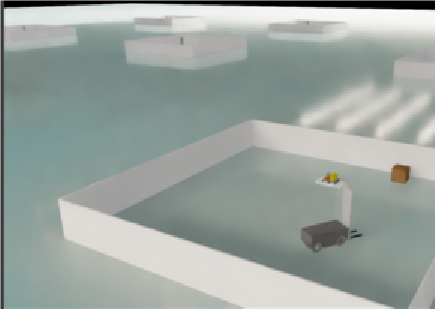}
         \subcaption{98386 lm, 2336 K, floor RGB (0.79, 0.85, 0.59)}\label{fig:dr_2}
     \end{minipage} \\
     \begin{minipage}[b]{0.45\linewidth}
          \centering
          \includegraphics[keepaspectratio, scale=0.5, trim= 0cm 0.2cm 0cm 0.6cm,clip]{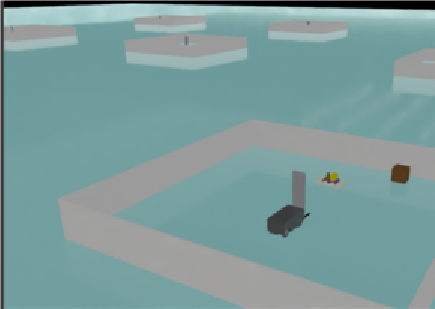}
          \subcaption{1875 lm, 5304 K, floor RGB (0.53, 1.00, 1.00)}\label{fig:dr_3}
      \end{minipage} 
     \begin{minipage}[b]{0.45\linewidth}
          \centering
          \includegraphics[keepaspectratio, scale=0.5, trim= 0cm 0.2cm 0cm 0.6cm,clip]{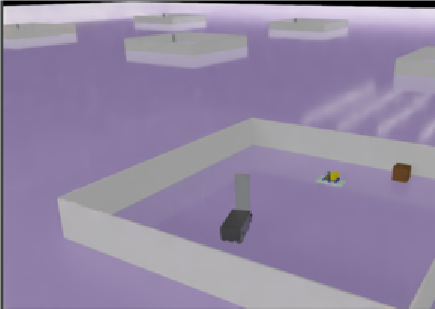}
          \subcaption{8563 lm, 6578 K, floor RGB(0.70, 0.55, 0.99)}\label{fig:dr_4}
      \end{minipage} 
     \caption{Example of domain randomization.}\label{fig:dr_sample}
\end{figure}

\begin{figure}[t]
	\centering
		\includegraphics[keepaspectratio, width=7.5cm]{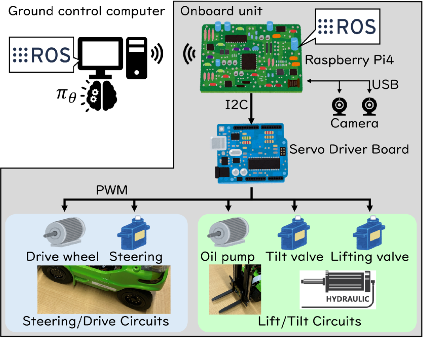}
	  \caption{Real forklift control system.}\label{fig:circuit}
\end{figure}

\begin{figure}[t]
\vspace*{1.4mm}
     \begin{minipage}[b]{0.5\linewidth}
         \centering
         \includegraphics[keepaspectratio, scale=0.22]{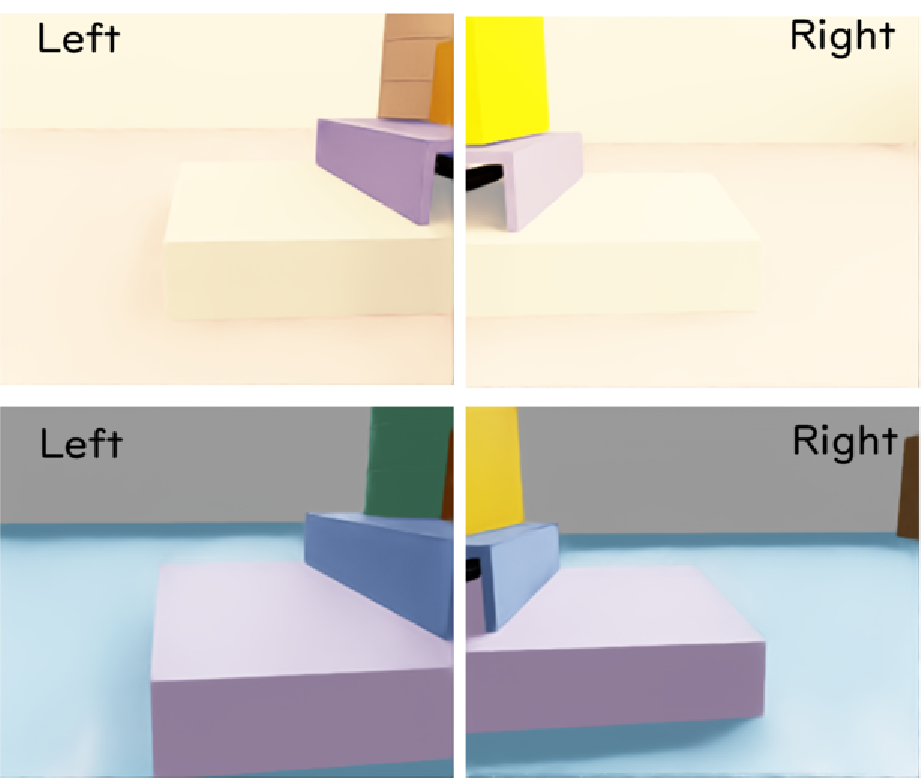}
         \subcaption{Success data}\label{fig:data_s}
     \end{minipage} 
     \begin{minipage}[b]{0.45\linewidth}
         \centering
         \includegraphics[keepaspectratio, scale=0.22]{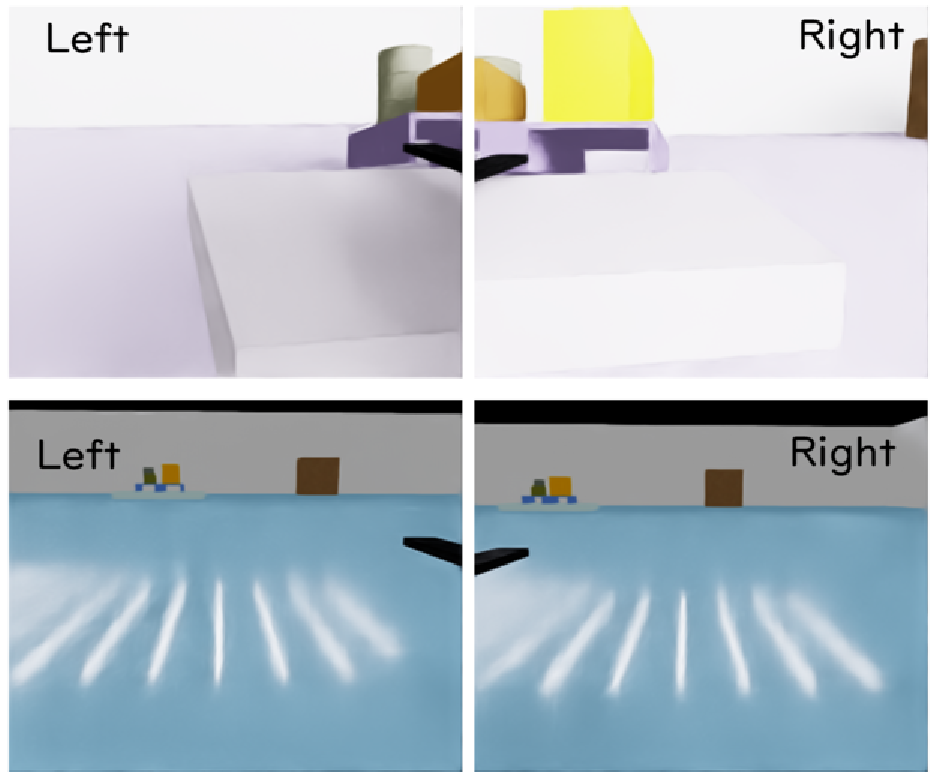}
         \subcaption{Fail data}\label{fig:data_f}
     \end{minipage}
     \caption{Dataset for decision policy.}\label{fig:data}
\end{figure}

\subsection{Digital and real environment} \label{sec:digi}
We constructed digital and real environments for the task involving approaching a pallet within a 1.8 m square space. 
The appearances of the digital and real environments are shown in Fig. \ref{fig:twin}.

The forklift was randomly placed within the green triangle shown in Fig.~\ref{fig:digi_env} and made to approach the pallet. 
This task relied on the onboard cameras mounted on the forklift for visual input. 
Because the forklift does not appear in the onboard cameras, we simplified its digital model to reduce the implementation complexity.
The objects in the digital environment were created using CAD data and specifications, without any real-world image capturing. 
As reflections are known to affect visual navigation in real environments negatively, fluorescent lights were modeled and placed as shown in Fig.~\ref{fig:ceiling} \cite{xue2024indoor}. 
The 1.8 m space was enclosed by white walls in the digital and real environments. 
The critical dimensions, namely the sizes of the pallet and forklift, are shown in Fig.~\ref{fig:fork_pallet}.
The specifications of the forklift are discussed in the following section.
Isaac Sim allowed for various randomizations within this digital environment. 
The information that was randomized in this study is listed in Table~\ref{tb:dr}. Examples of the floor and lighting randomization are shown in Fig.~\ref{fig:dr_sample}.

\subsection{Real forklift} \label{sec:real_fork}
For the real-world environment, we modified a 1/14-scale forklift from LESU, which is a China-based manufacturer \cite{lesu2024}. 
The dimensions of this forklift were 100 $\times$ 310 mm, and the forks measured 75 $\times$ 10 mm. 
Similar to counterbalance forklifts, this forklift featured front-wheel drive and rear-wheel steering.
In addition, the lift and tilt functions of the fork were powered by hydraulics.

All computers installed on this forklift were replaced with a Raspberry Pi 4 single-board computer to implement the ROS.
Raspberry Pi received action $\bm{a}$, calculated by the ground control computer based on the approach policy $\pi_{\bm{\theta}}(\bm{a} | \bm{o})$, and transmitted these commands to the respective servos and motors via the servo driver board, as shown in Fig.~\ref{fig:circuit}. 
Two USB cameras were connected to the Raspberry Pi; the captured images were sent to the ground control computer as observation. 
The cameras were mounted on both sides of the forklift, as shown in Fig.~\ref{fig:fork_pallet}, and were equipped with OV5640 sensors and 60-degree field-of-view lenses.

Velocity information was obtained using a motion capture system. 
Note that only velocity information was used in our method. The position data were only used for experimental evaluation.
All control processes were executed at 15 Hz.

\subsection{Digital forklift} \label{sec:digi_fork}
The forklift in the digital environment was implemented using a simplified design, as mentioned in Section~\ref{sec:digi}.
The drivetrain components were operated using the Articulations-based controllers of Isaac Sim. 
The parameters of these controllers were configured to match the speed response of the real forklift to the command inputs closely.

\subsection{DRL on Isaac Sim} \label{sec:IsaacSim}
We used ``OmniIsaacGymEnvs,'' which is a parallel learning framework that is based on ``rl\_games'' and is available in Isaac Sim \cite{RN56, OmniEnvs}. 
This tool enables the construction of multiple environments on the same ground plane, thereby improving the sampling efficiency. 
The randomization of the colors for the pallets and other objects was handled independently for each environment for domain randomization. 
We synchronized the task resets and shared the values for lighting and floor color randomization across all environments to simplify the implementation.

\subsection{Decision policy for loading} \label{sec:load}
To perform loading, verifying that the approach to the pallet has been successful is crucial.
In this study, we designed a decision policy that activates when the forklift is stationary. 
The decision policy determines whether to lift the forks after the forklift approaches the pallet using the approach policy obtained through DRL.
This policy was trained using binary classification through supervised learning. 
The training dataset was generated by collecting camera images from successful and unsuccessful attempts within the Isaac Sim environment, as shown in Fig.~\ref{fig:data}. 
Consequently, no real-world data were used to create this dataset.

\section{Real demonstration} \label{sec:demo}
This section presents the results of deploying our method in a real environment.
As our method employs zero-shot sim2real, the policies trained in the digital environment were transferred directly to the real environment.

\subsection{Setup} \label{sec:demo_set}
We trained the policies in the digital environment described in Section \ref{sec:sysfork} using the method outlined in Section \ref{sec:RL}.
A ground control computer with 96 GB of RAM, an Intel{\textsuperscript{\tiny\textregistered}} Xeon{\textsuperscript{\texttrademark}} Gold 6146 3.2 GHz processor, and an NVIDIA RTX A6000 was used. 
The task was performed by positioning the forklift within a triangular area, similar to that during the training in the digital environment, as shown in Fig.~\ref{fig:twin}. 
In this demonstration, we repeatedly positioned the forklift at an initial position, directed it to approach the pallet, and manually returned it to a new starting position after each task.

The decision policy for determining whether to lift the forks was triggered 3 s after the forklift came to a stop. 
In the case of a successful approach decision, the forklift followed a sequence of actions: lifting the forks, reversing, and lowering the forks to unload the pallet. 
Successful completion of the task was dependent on executing the sequence without dropping the pallet load.

We compared the results of our method with those of human-operated trials to demonstrate the difficulty of the task. 
Human operators controlled the forklift using a joystick, relying on the same camera images and velocity information as the approach policy. 
The operators were provided with a brief explanation and a 3-min practice session before the experiment.
As hydraulic operation requires experience, the experiment was considered successful if the operator could insert the forks at least two-thirds into the pallet.

\begin{table}[t]
\centering
\caption{Demonstration results}
\begin{tabular}{|c|c|c|c|c|}
\hline
  & Ours & Man A & Man B & Man C \\ \hline
Success rate (\%)  &  60  & 50   & 40   & 90   \\ \hline
Time (s)  & 6.5   & 7.7   & 17.2   & 24.5  \\ \hline
\end{tabular}
\label{tb:result}
\end{table}

\begin{figure}[t]
\vspace*{1.4mm}
	\centering
		\includegraphics[keepaspectratio, width=7.5cm]{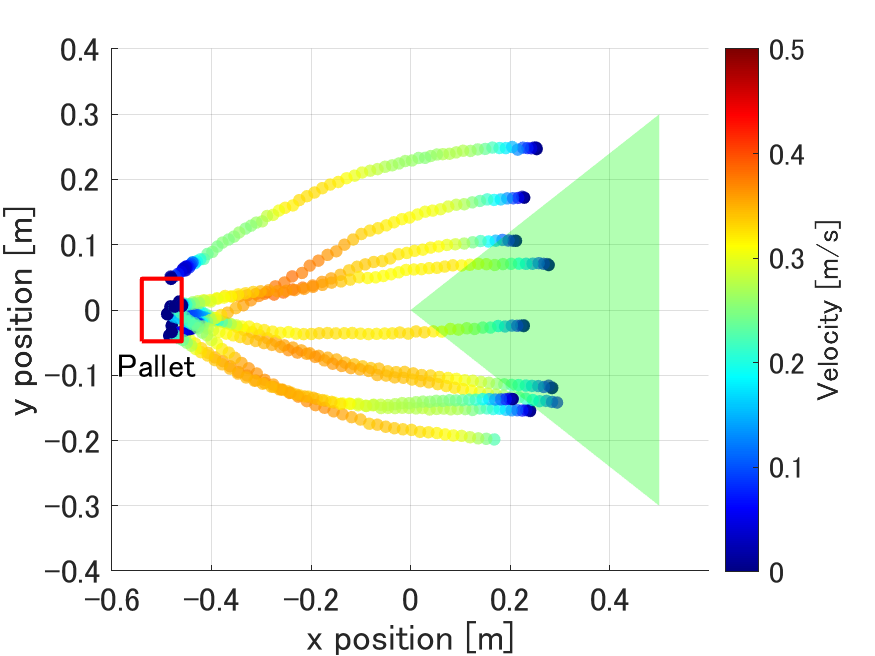}
	  \caption{Trajectory and velocity of the center of the fork. The green triangle indicates the range of randomized initial positions during digital training. As the trajectory represents the path of the fork, the center of the forklift body is positioned 0.2 m further back. }\label{fig:result}
\end{figure}

\begin{figure}[t]
	\centering
		\includegraphics[keepaspectratio, width=7.2cm, trim= 0.0cm 0.7cm 0cm 1.0cm,clip]{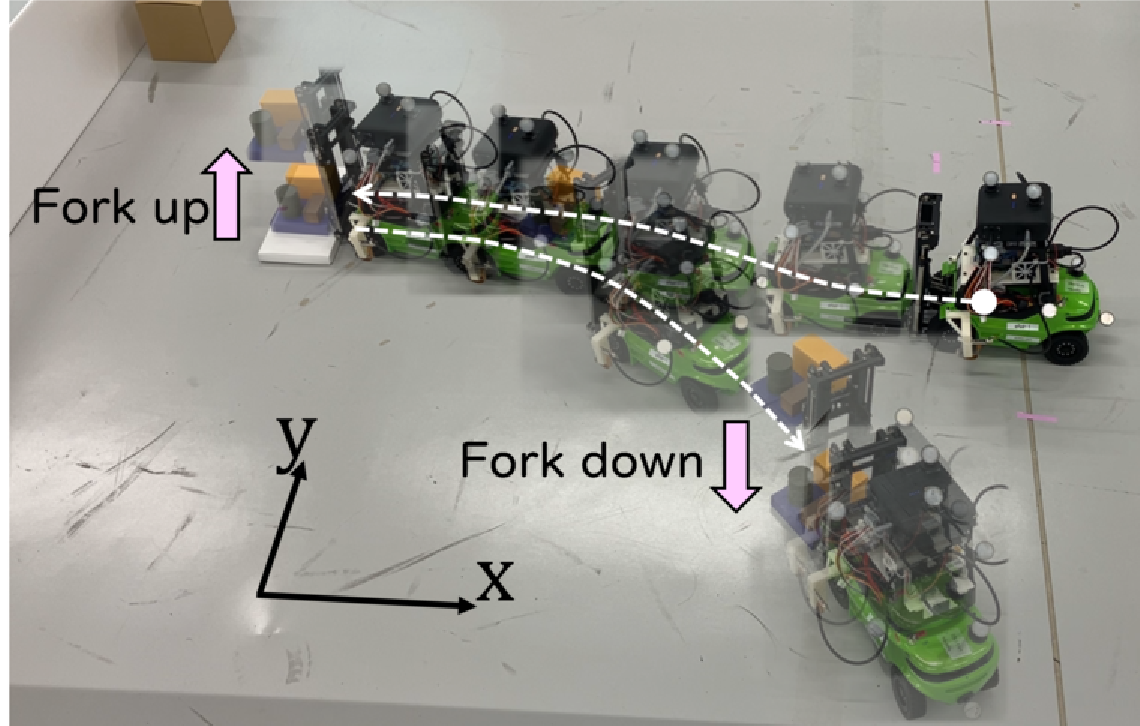}
	  \caption{Demonstration of forklift task.}\label{fig:snap}
\end{figure}

\subsection{Results}\label{sec:reslut}
We performed the task 10 times with our learned policies and the three human operators.
The success rates and average times of the pallet approach required for success are shown in Table~\ref{tb:result}.
The average success rate for human operators was 60\% and the completion time was 16.5 s. 
Although Man C achieved a success rate of 90\%, he employed a cautious driving technique, stopping repeatedly and checking the camera images frequently. 
This indicates that the task was not straightforward.
Our method achieved a 60\% success rate in the pallet approach task at the highest speed.
Figure~\ref{fig:result} shows the position and velocity of the forklift forks with our method during the 10 trials based on the motion capture system.
This result suggests that our approach policy learned efficient behavior, accelerating initially and decelerating near the pallet.
Furthermore, the success rate of the loading decisions was 90\%. 
A decision error occurred once when the forks were inserted only on one side.
Snapshots of a successful attempt are presented in Fig.~\ref{fig:snap}. 

Our method could produce a controller for forklift tasks using only visual and velocity data, similar to human operation, although the performance was not ideal.
Furthermore, the experiment was safe and easy to conduct.

\section{Conclusions and future work} \label{sec:concl}
We have proposed a learning system for a counterbalance forklift to perform pallet loading tasks using only cameras and velocity data. 
In addition, we developed a 1/14-scale forklift with the same configuration as the 1/1-scale one. 
The forklift allows us to validate the trained policies for safety. 
Moreover, our method eliminates reliance on real-world data by utilizing a photorealistic environment and domain randomization.
Our approach policy achieved a 60\% success rate through efficient behavior in loading tasks in real-world experiments.
In addition, the forklift could perform loading tasks based on the success or failure of its approach to the pallet using a decision policy learned from a dataset constructed with a digital environment.
These policies were implemented through zero-shot sim2real, requiring no heuristic adjustments.
Based on these results, this study presented a practical method for applying learning-based control to counterbalance forklifts, offering significant potential to advance automation across various industries.

Future work will extend our training approach and environment to accommodate more diverse scenarios. 
In addition, we will incorporate insights from existing studies to improve the accuracy further and validate these results on a full-scale forklift.

\clearpage
\bibliographystyle{IEEEtran}
\bibliography{Fork_ICRA2025}

\end{document}